\documentclass[runningheads]{llncs}
\usepackage[T1]{fontenc}
\usepackage{indentfirst}
\usepackage{graphicx}
\usepackage{amsmath}
\usepackage{txfonts}
\usepackage{subfig}
\begin{document}
\title{Positive-Sum Fairness: Leveraging Demographic Attributes to Achieve Fair AI Outcomes Without Sacrificing Group Gains}
\titlerunning{Positive-Sum Fairness}
%

\author{Samia Belhadj$^{*,}$\inst{1} \and
Sanguk Park$^{*,}$\inst{1}\orcidID{0009-0005-0538-5522} \and 
Ambika Seth\inst{1} \and
Hesham Dar\inst{1}\orcidID{0009-0003-6458-2097} \and
Thijs Kooi\inst{1}\orcidID{0000-0001-7701-7837}}

%
\authorrunning{S. Belhadj \& S. Park et al.}
%
\institute{
Lunit Inc., Seoul, Republic of Korea
\\
\email{\{samia.belhadj, tony.superb, ambika.seth, heshamdar, tkooi\}@lunit.io}}

\maketitle 
\def\thefootnote{*}\footnotetext{These authors contributed equally to this work}

\begin{abstract}
Fairness in medical AI is increasingly recognized as a crucial aspect of healthcare delivery. While most of the prior work done on fairness emphasizes the importance of equal performance, we argue that decreases in fairness can be either harmful or non-harmful, depending on the type of change and how sensitive attributes are used. To this end, we introduce the notion of positive-sum fairness, which states that an increase in performance that results in a larger group disparity is acceptable as long as it does not come at the cost of individual subgroup performance. This allows sensitive attributes correlated with the disease to be used to increase performance without compromising on fairness.  

We illustrate this idea by comparing four CNN models that make different use of the race attribute in the training phase. The results show that removing all demographic encodings from the images helps close the gap in performance between the different subgroups, whereas leveraging the race attribute as a model's input increases the overall performance while widening the disparities between subgroups. These larger gaps are then put in perspective of the collective benefit through our notion of positive-sum fairness to distinguish harmful from non harmful disparities.

\end{abstract}
\section{Introduction}
Medical imaging plays a critical role in diagnosis, treatment planning, and monitoring patient progress. However, the reliability of medical imaging algorithms is not uniformly distributed across different demographic groups, raising concerns about fairness and potential biases in the results. Fairness in medical imaging most often refers to the equitable treatment of patients from diverse demographic backgrounds, regardless of their gender, race, ethnicity, or other characteristics sensitive to discrimination \cite{lara2022addressing,xu2023fairness}.

This equitable treatment is often interpreted as a similar performance across different demographic subgroups. When applied to domains like credit card scoring or AI-powered recruiting, ignoring all sensitive attributes and prioritizing a similar performance across the different demographic subgroups is an acceptable approach. However, in the medical field, demographic attributes are important clinical factors which radiologists and clinicians often take into consideration as they can have a strong impact on their diagnoses and can guide them to consider specific tests or treatments based on the patient's demographic profile. The prevalence of diseases can be correlated to demographic attributes. For example, studies have shown that breast cancer has a higher incidence among Ashkenazi Jewish women \cite{Warner1999-gh,Rubinstein2004-iq}. And, due to historical and social disparities as well as different physiological features across demographic subgroups, the difficulty level of medical tasks is not uniformly distributed. For this reason, even collecting more or more diverse data does not necessarily produce equal performance across demographic subgroups as the best achievable result is not the same for each of them \cite{Petersen2023}. In a domain where each improvement can save lives, it is hard to disregard the benefit of the population as a whole for the sake of decreasing the disparities between subgroups. 

Petersen et al. \cite{petersen2024demographicallyinvariantmodelsrepresentations} examined various types of demographic invariance in medical imaging AI, highlighting why they can be undesirable and stressing the need for better fairness assessments and mitigation techniques in this field. Several fairness measures suffer from degradation in the overall performance by penalizing the performance of an AI system for groups that it performs better on, in order to achieve parity with groups it performs worse on, which is referred to as “levelling down” \cite{mittelstadt2023unfairnessfairmachinelearning}. While we are aware of papers suggesting training methods which aim to maximize the benefit of each subgroup (Berk Ustun \cite{pmlr-v97-ustun19a}, for instance, suggested debiasing methods following the ethical principles of beneficence (“do the best”) and non-maleficence (“do not harm”) \cite{Varkey2021-lu} in regards to fairness), and methods which improve fairness by understanding and mitigating the demographic encodings present in images \cite{yang2023limits,Brown2023-kk}, we could not find any fairness evaluation framework or definition which allows to compare different models from the prism of harmful and non harmful disparities.

\begin{figure}
    \centering
    \begin{minipage}{.3\linewidth}
    \centering
    \subfloat[]{\includegraphics[height=3cm]{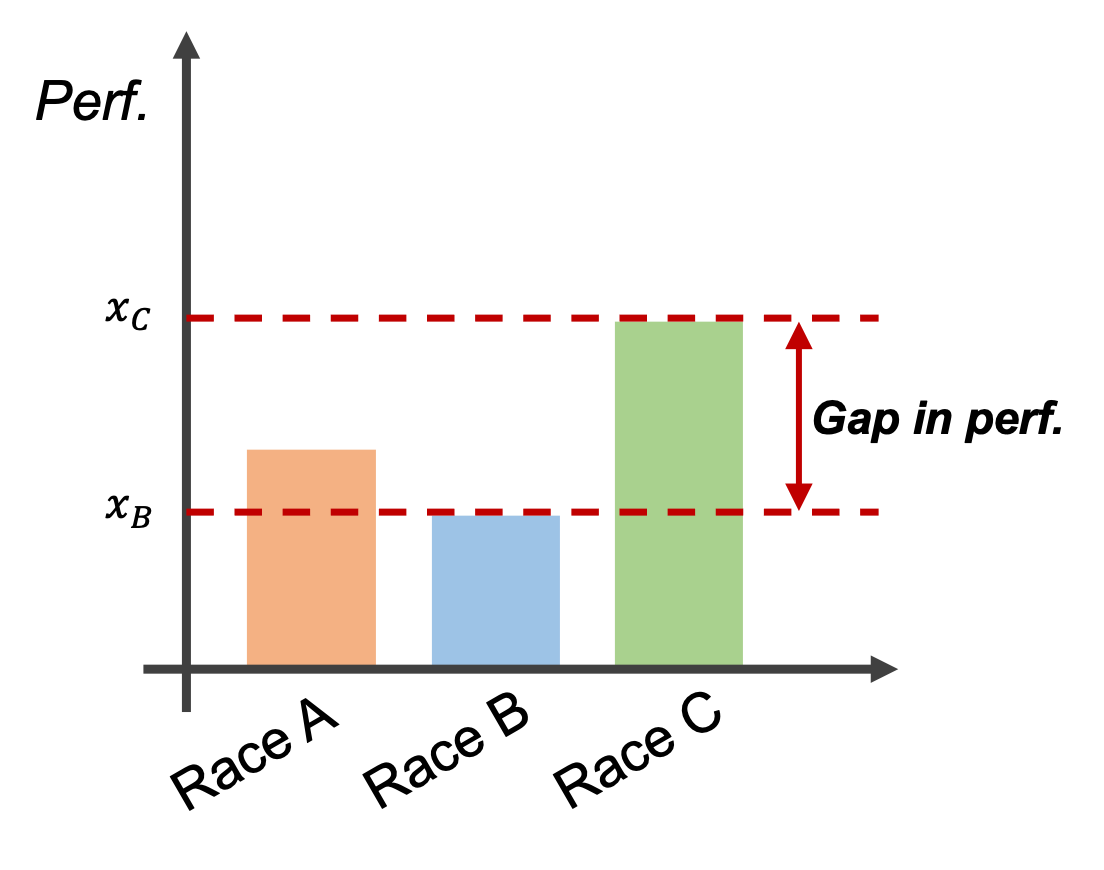}}
    \end{minipage}
    \begin{minipage}{.3\linewidth}
    \centering
    \subfloat[]{\includegraphics[height=3cm]{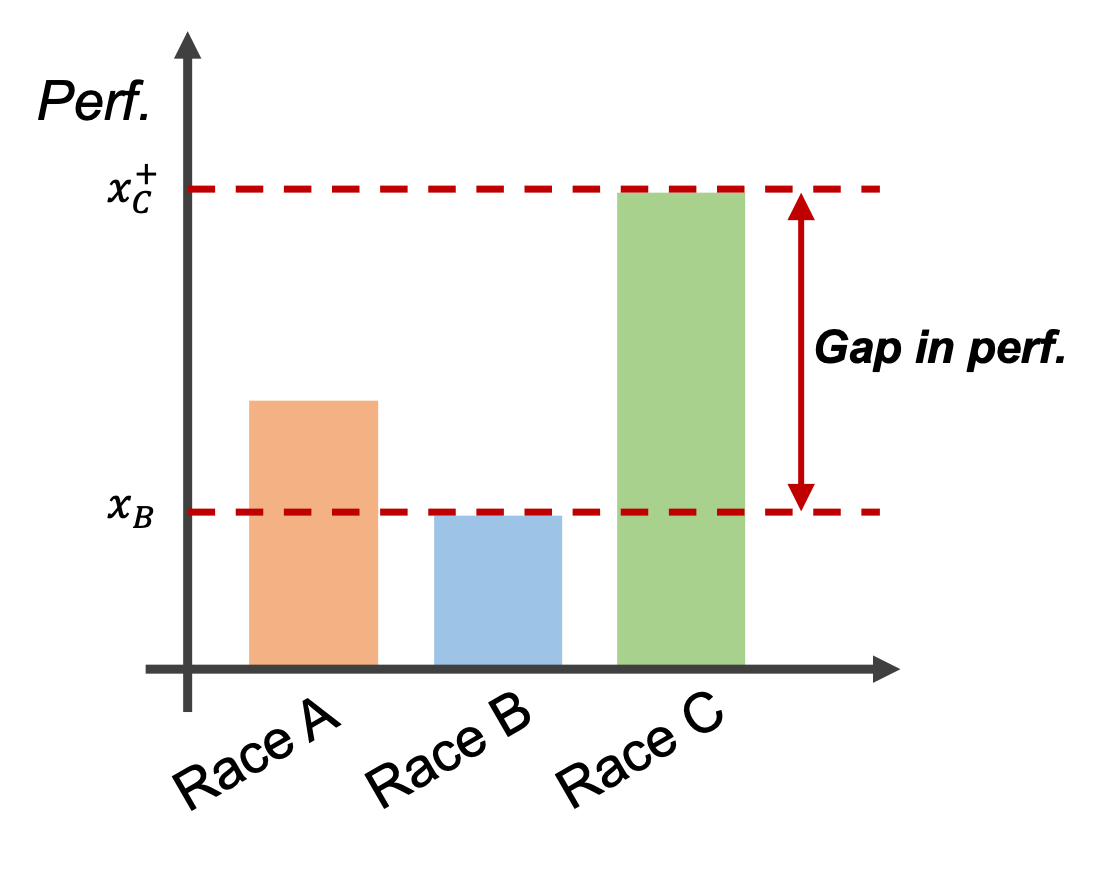}}
    \end{minipage}
    \begin{minipage}{.3\linewidth}
    \centering
    \subfloat[]{\includegraphics[height=3cm]{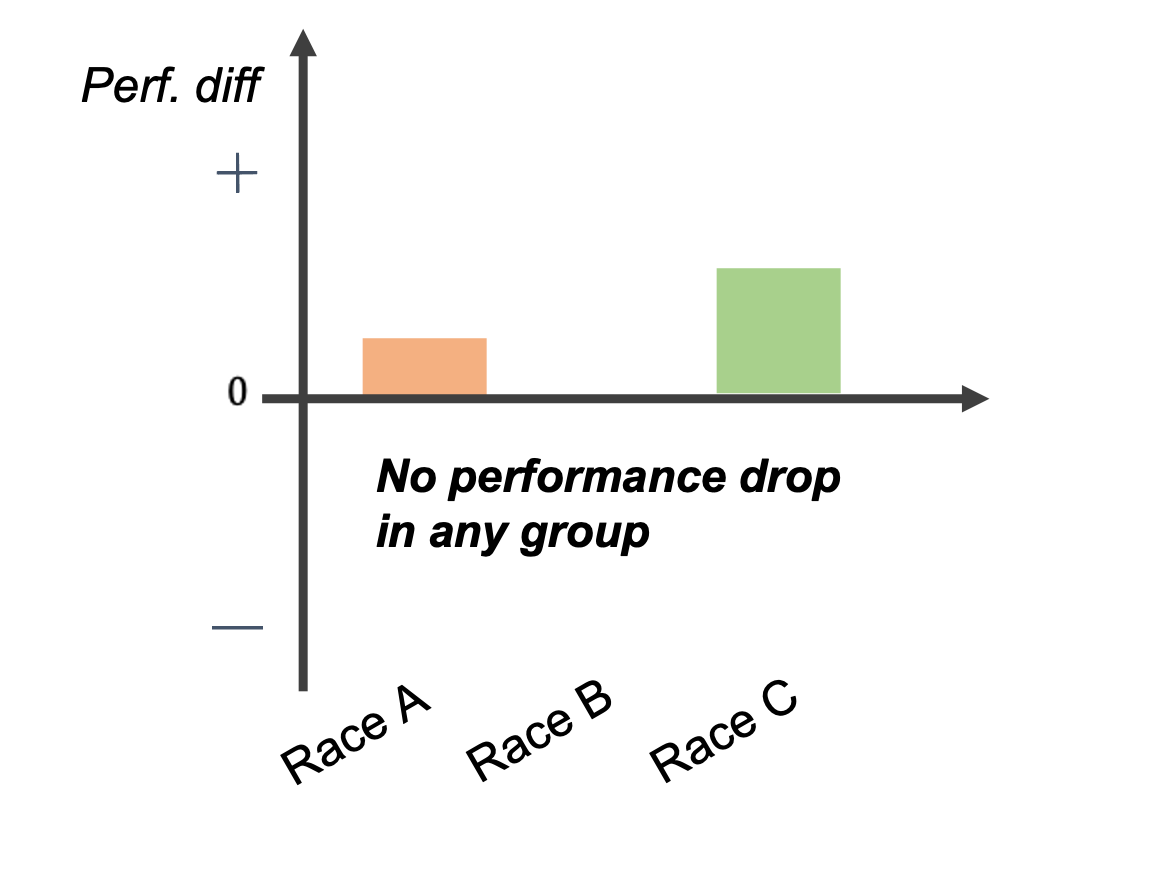}}
    \end{minipage}
    \caption{We investigate fairness of AI models and introduce the concept of 'positive-sum fairness' to differentiate {\it harmful} and {\it non-harmful} disparities. 
    Graph a) shows the performance of an initial model per protected groups. b) shows the performance of an updated model with a higher overall performance but a lower fairness, under its standard definition, as indicated by the larger difference between the most and least advantaged groups and therefore could be rejected on the basis of fairness. c)
    shows the same updated model as b) however it shows the performance difference per group compared to the initial model. In this positive-sum framing we see that none of the groups had a reduction in performance and therefore the increased performance in Race C did not come at the cost of performance in any other group.}
    \label{fig:fairness_definition_in_paper}
\end{figure}

We, therefore, introduce the notion of {\it positive-sum fairness}: when looking at a situation where we have an initial model and are looking at the trade-off between fairness and performance while trying to improve it, inequitable performance can be acceptable as long as it does not come at the expense of other subgroups and allows a higher overall performance to be achieved. Specifically, we argue that differences in performance can be {\it harmful} and {\it non-harmful}. 
We consider a disparity harmful if it comes at the cost of the overall performance {\it or} if improving the overall performance is achieved by decreasing performance on any protected subgroup. A difference in performance across protected subgroups is considered non-harmful if, by improving an AI system's performance, we exacerbate the disparities between subgroups without negatively impacting any specific subgroup.
This main idea is summarized in figure \ref{fig:fairness_definition_in_paper}.

We compare the positive-sum fairness framework with a more traditional group fairness definition, which is the largest disparity in performance across subgroups. We show that some models, while increasing this disparity, actually improve the performance of each subgroup individually and that other models which decrease the disparity ("improving fairness" from a classic point of view) are harming some subgroups to achieve it.

\section{Related work}
Bias is commonly identified in medical image analysis applications \cite{xu2023fairness,zong2023medfair}. For instance \cite{eam2022fairnessrelated}, a CNN trained on brain MRI resulted in a significant difference between ethnicities. Seyyed-Kalantari et al. \cite{seyyed21} observed that minorities received higher rates of algorithmic underdiagnosis. Zong et al. \cite{zong2023medfair} assessed bias mitigation algorithms in- and out-of-distribution settings. The experiments demonstrated the wide existence of bias in AI-based medical imaging classifiers and none of the bias mitigation algorithms was able to prevent this.

Different definitions of fairness are used:
\begin{itemize}
\item {\bf Individual fairness} \cite{mukherjee2020simple} requires that similar individuals should be treated equally and thus have similar predictions. For example, a model should have comparable diagnosis on two similar X-Ray images.
\item {\bf Group fairness} requires equal performance on sub-groups divided based on sensitive attributes (e.g., race, sex, and age). Common group fairness metrics are demographic parity \cite{feldman2015certifying}, equal odds \cite{hardt2016equality} and predictive rate parity or sufficiency \cite{Lee2021FairSC}. 
\item {\bf Minimax fairness} \cite{diana2021minimax} seeks to ensure that the worst-off group is treated as fairly as possible, reducing the most severe negative impacts of a decision or system.\end{itemize}
These definitions have pros and cons \cite{Verma2018FairnessDE}. Individual fairness relies on the choice of the distance metric, which requires expert input. In minimax fairness, the ideal solution is difficult to compute and the degree of unfairness relies heavily on the choice of the set of models. Group fairness metrics are easy to implement and understand, but are not always adapted to the problem nor compatible with one another \cite{berk2017fairness,kleinberg2016inherent}. And even though prior work has broadened the group fairness notion by adding other normative choices than strict equality \cite{baumann2023distributivejusticefoundationalpremise}, none of the proposed metrics prevent the harm that could be brought to each subgroup's performance individually or to the whole population's benefit.

As mentioned in the introduction, similarly to \cite{mittelstadt2023unfairnessfairmachinelearning,pmlr-v97-ustun19a,Petersen2023,petersen2024demographicallyinvariantmodelsrepresentations}, we believe that medical AI is different from other domains in that each improvement can save lives. Therefore, increasing disparities to achieve the best performance possible for each demographic subgroup and for the population as a whole could be justified. Previous research has shown that images themselves could carry demographic encodings \cite{Glocker2023-xj,Gichoya2022-qm}. E.g., Yang et al. \cite{yang2023limits} investigate the utilization of demographic encodings by analyzing the use of demographic shortcuts for disease classification. Two papers \cite{custers2016using,9308585} examine the relevance of explicitly using sensitive attributes in fair classification systems for non-medical problems. They compare different models which leverage sensitive attributes with a model which is not trained on any sensitive attribute.

\section{Methods}

\subsection{Positive-sum fairness}
\label{sec:meth_positive_sum_fairness}
We introduce the principle of positive-sum fairness, which analyzes fairness from the prism of {\it harmful} and {\it non harmful} disparities. When looking at changes in model performance and disparities between protected subgroups, there are several explanations for a gap in performance between the most and least advantaged subgroups: 
\begin{itemize}
\item The most advantaged group's performance improved while others' stayed the same,
\item All subgroups' performance improved but one of them increased more than others,
\item The most discriminated group's performance decreased while others' stayed the same,
\item All subgroups' performance decreased, but one of them decreased more than the others, etc.
\end{itemize}
The first two would not be considered harmful as they allow to improve the general performance without harming any of the subgroups, thus achieving a collective benefit.

\subsubsection{Definition}

Positive-sum fairness is a fairness evaluation framework where the goal is to find solutions that increase the overall benefit for all parties together while trying to ensure no one is worse off and ideally, everyone is better off. It looks at the situation where we have an initial model and are looking at the trade-off between fairness and performance when trying to improve the model. Unlike other fairness definitions which aim to minimize the disparity between subgroups or maximize the worst performance among subgroups, positive-sum fairness tries to avoid gains to a group which come \textit{at the expense} of another group while maintaining the overall performance.

Let us assume that we compare $N$ models $\{M\}^{N}_{i=1}$ to a baseline $M_{baseline}$ on $K$ demographic subgroups. And let us consider $measure(M)$ as the metric that measures the performance of a model $M$. Following the positive-sum fairness definition, selecting the best model is equivalent to finding the best trade-off between: 
\begin{itemize}
\item maximizing the performance gain: $\max_{1 \leq i \leq N} measure(M_{i}) - measure(M_{baseline})$
\item maximizing the smallest performance gain across the subgroups :\newline 
$\max_{1 \leq i \leq N} (\min_{1 \leq k \leq K}  measure(M_{i})(group_{k}) - measure(M_{baseline})(group_{k}))$
\end{itemize}

Depending on the task, one could set hard constraints like ensuring there is no performance loss for any subgroup (aka the selected model $M_{i}$ should ensure that $\min_{1 \leq k \leq K}  measure(M_{i})(group_{k}) - measure(M_{baseline})(group_{k}) \geq 0 $) and  the overall performance is improved (aka the selected model $M_{i}$ should ensure that $measure(M_{i}) - measure(M_{baseline}) \geq 0 $) or find the most relevant trade-off between the two optimization problems.

\subsection{Application}

To put this fairness notion into practice and show the difference with traditional group fairness, we compare three models which use sensitive attributes to a baseline model. The way sensitive attributes are used by the model is known to have an impact on the fairness and performance of the model \cite{Brown2023-kk,yang2023limits,custers2016using,9308585}. Therefore, we make use of models that explicitly include sensitive attributes, or conversely, remove any demographic encoding from the input data.

\begin{figure}[h]
\includegraphics[scale=0.3]{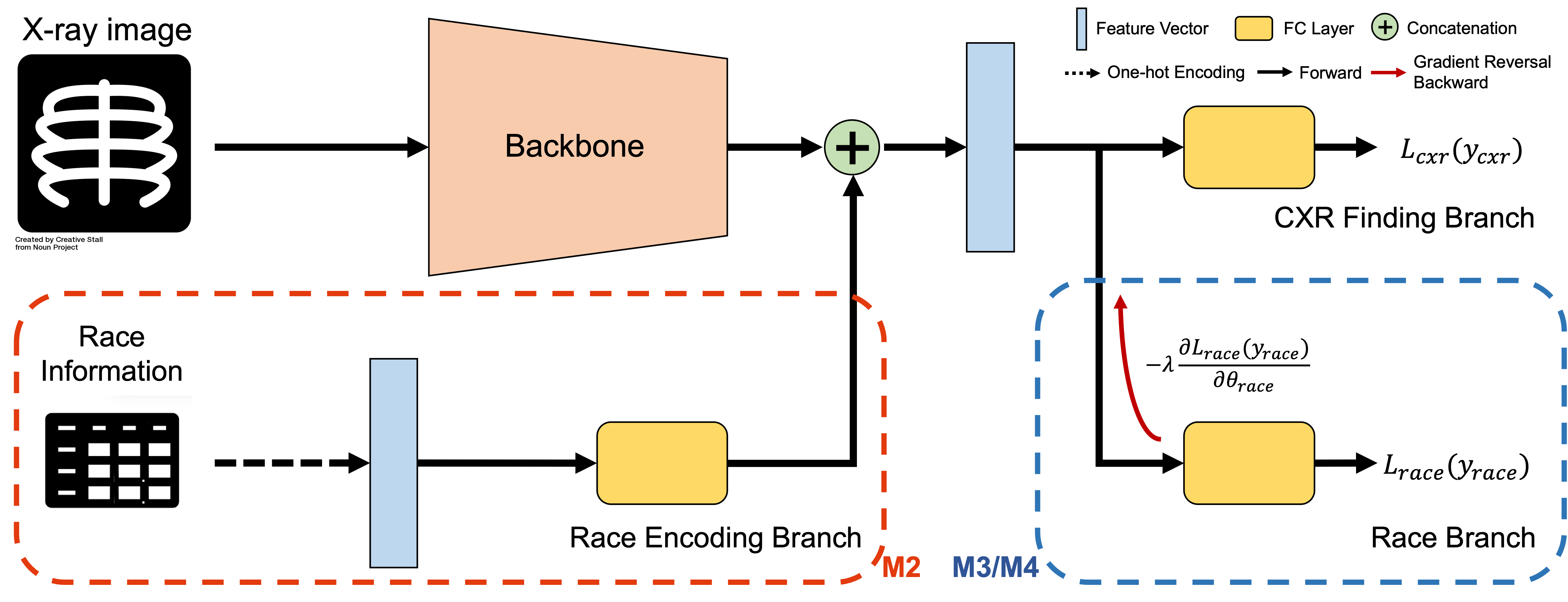}
\caption{To investigate the effect of sensitive attributes on performance and fairness, we evaluate four different model architectures, denoted M1, M2, M3 and M4. M1, the baseline, has a backbone and classification. M2 has a race encoding branch to learn race-encoded features directly from metadata. M3 and M4 have an additional race branch to predict the race group which is implicitly encoded in the image, from the image features. The difference between M3 and M4 is that we add a gradient reversal layer before the race branch.}
\label{fig:model_archs}
\end{figure}

The four models are trained on a multi-label classification problem of findings in chest radiography (CXR). In all settings, a Densenet-121 \cite{huang2018densely} backbone is used, which was empirically determined to give the best performance for this problem. The exact model architectures are shown in figure \ref{fig:model_archs} and described below: 

\begin{itemize}
\item \textbf{M1}: a baseline classifier using the images as input and trained to predict the targeted CXR findings associated to our dataset. The model comprises a backbone to extract the image features and a finding branch consisting of a fully connected layer and a binary cross entropy loss for each finding. 
\item \textbf{M2}: a classifier using both the images and race features as input. The race information comes in the form of a categorical variable, which we convert to a one-hot vector and feed to a fully-connected layer. We concatenate the features from the fully connected layer and the image features before forwarding to finding branch. The model is trained end-to-end.
\item \textbf{M3}: a classifier using the images as input only, but trained to predict image findings as well as the race group (i.e. this model aims to exploit the race encodings present in the images). For this model, we modify the final layer of the baseline classifier by adapting the loss function to optimize the two tasks: CXR findings and race group. We also transform race information to one-hot encoded vector to apply multi-class loss. The race classification branch is made of a fully-connected layer and a cross entropy loss function. The final loss is calculated by adding finding loss and the race loss with a loss weight $\lambda$.
\[
L(y_{cxr}, y_{race}) = L(y_{cxr}) + \lambda L(y_{race})
\]
\item \textbf{M4}: a classifier using the images as input, trained to predict image findings, while minimizing the use of race information encoded in the image. For this model, we implement the gradient reversal technique described in \cite{raff2018gradient}. We apply the gradient reversal layer before the race branch. 
\end{itemize}

\section{Experiments}
\label{sec:experiments}

\subsubsection{Data}
We use chest radiographs from MIMIC-CXR-JPG \cite{johnson2019mimiccxrjpg,inbook}. The dataset has annotations for 14 findings. However, we focus on lung lesions, pneumonia, pleural effusion and consolidation as the diseases associated with these findings have been shown to be correlated with ethnicity \cite{Burton2010-rf,Joseph2020-za,Shi2022-ub}. We use only frontal images and split the dataset into training, validation, and test sets on a patient level. In total, 237,972, 1,959, and 3,403 images are used for training, validation, and testing, respectively.
\subsubsection{Sensitive attributes}
We define the protected subgroups based on the self-reported race from MIMIC-IV \cite{Johnson2023-td,Johnson2023-ex} and split it into five groups: White, African-American, Latino, Asian, others.
\subsubsection{Model training}
We train our 4 models to predict all 14 CXR findings and a race group. We initialize a DenseNet-121 backbone with pre-trained weights from ImageNet \cite{russakovsky2015imagenet}. The images are resized to 256 $\times$ 256, and augmented using random rotation from [-15,15] degree range and random horizontal flip. We conduct the experiments with 8 V100 NVIDIA GPU. AdamW \cite{loshchilov2019decoupled} is used with an initial learning rate of 0.002 which is updated using the cosine annealing warm up \cite{loshchilov2017sgdr} scheduler.
\subsubsection{Evaluation}
We compare the four models by general performance and fairness across the protected subgroups. The general performance is assessed using the Area under the ROC curve (AUROC) score and the traditional group fairness metric used to compare with positive-sum fairness is expressed by (1 - largest disparity between protected subgroups in terms of AUROC) \cite{lee2023fairness}. We use the AUROC mean and confidence bounds generated using bootstrapping with 300 samples \cite{efron1987better}. We do not consider protected subgroups which have less than 5 positive cases or less than 5 negative cases as this results in poor estimates of performance.

\subsection{Initial results}
\begin{figure}
\centering
\subfloat[Traditional group fairness vs performance framework\label{subfig-1:dummy}]{%
    \centering
    \includegraphics[width=0.55\textwidth]{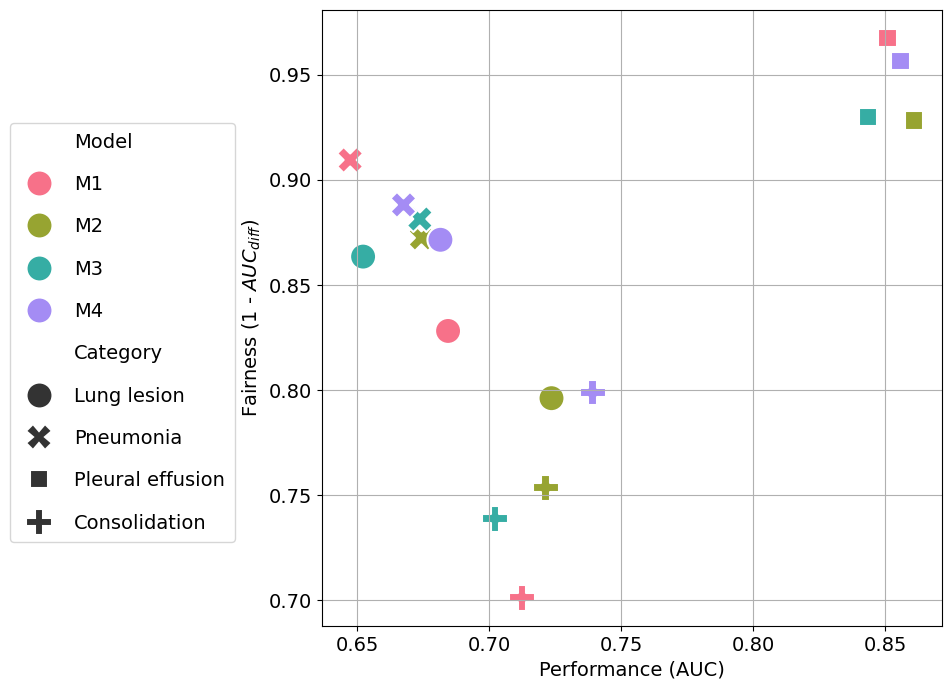}
    \label{fig: fig3}
    }
\subfloat[Positive-sum fairness framework\label{subfig-1:dummy}]{%
    \centering
    \includegraphics[width=0.4\textwidth]{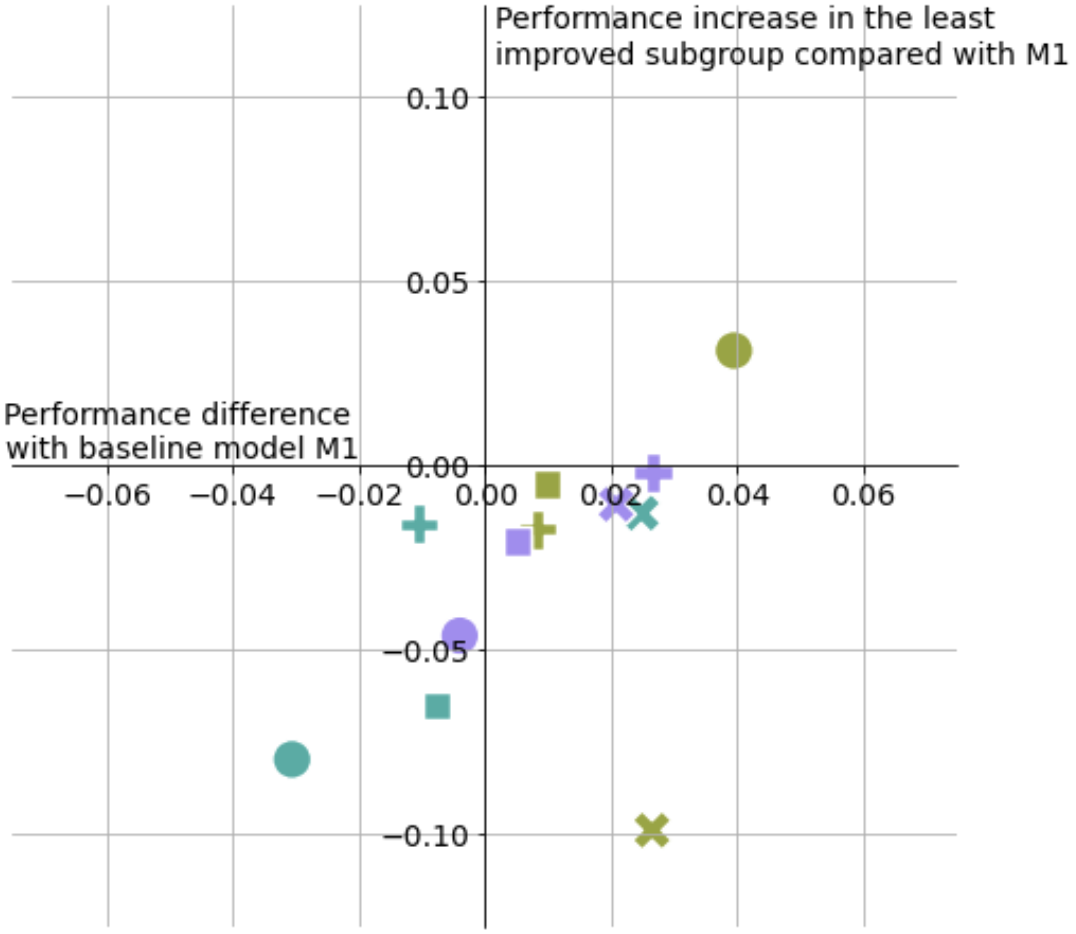}
    \label{fig: fig5}
    }
\caption{We put in parallel 2 different fairness vs performance frameworks: in figure (a), we compute both the performance (AUROC) and fairness (as 1 - the difference in AUROC between the most and least advantaged groups) of the 4 models per lesion. And in figure (b),  we show, the difference in overall performance and in performance per protected subgroup between the 3 improved classifiers and the baseline M1. The x axis compares the performance of each improved classifier with the baseline and the y axis shows whether at least one protected subgroup has been harmed by the modifications brought to the baseline classifier.}
\end{figure}

According to traditional group fairness, in assessing the results of the four models shown in figure \ref{fig: fig3} one could conclude that:

{\bf M2 improves the overall performance}
Our results show that M2 outperforms M1 in terms of AUROC. This is in line with our expectation as we are providing an additional relevant medical feature for the model to learn from. This better performance comes with a larger gap in AUROC between the most advantaged and most discriminated races, in other words less fairness from a traditional point of view. But this larger disparity is not necessarily {\it harmful} according to the positive-sum fairness as we will discuss it in the next section.

{\bf M4 improves the fairness}
M4 improves fairness for lung lesions and consolidations, while performing similar for pneumonia and pleural effusion. The improved fairness is likely due to the gradient reversal layer, which removes race information from the image and prevents the model from exploiting any demographic shortcut.

{\bf No clear pattern for M3}
The results for M3 are less consistent. Its performance is lower than the baseline except for pneumonia and its fairness measurement is sometimes higher and other times lower than the baseline's. If the baseline model exploited demographic encodings present in the images to generate shortcuts, training M3 to maximize the race prediction might have intensified the impact of these shortcuts.

\subsection{Positive-sum fairness}
We now apply the notion of positive-sum fairness, defined in section \ref{sec:meth_positive_sum_fairness} and reframe the fairness vs performance problem as shown in  \ref{fig: fig5}.
Here, the x-axis represents the difference in performance between each improved classifier and the baseline ($AUROC(M_{i}) - AUROC(M_{1})$) and the y-axis shows the performance increase (or decrease) for the least improved subgroup ($\min_{1 \leq k \leq K}  AUROC(M_{i})(race_{k}) \\ - AUROC(M_{1})(race_{k})$). A negative value indicates that the model performs worse for the given subgroup.

Any classifier which has the exact same overall performance and exact same performance per protected subgroup (race) as the baseline, would be at coordinate (0,0). Any classifier that has a negative x-coordinate, would have a lower general performance than M1 and any classifier that has a negative y-coordinate would have at least one protected subgroup with a lower AUROC than M1 (at least one subgroup negatively impacted by the changes brought to the baseline model). 

For lung lesions, figure \ref{fig: fig5} shows that M2 appears in the positive side of the x and y axes, meaning that the performance was improved without harming any subgroup's performance. And this even though the figure \ref{fig: fig3} shows a decrease in fairness (larger disparity between the most advantaged and least advantaged subgroups) for M2 compared with M1. This matches the previous conclusion that the larger performance gap between protected subgroups for M2 compared with M1 cannot be considered harmful as every protected subgroup's performance was individually increased.

On the other hand, for lung lesions, model M4 improved fairness (smaller disparity between the most advantaged and least advantaged subgroups) as shown in the figure \ref{fig: fig3}. However, the figure \ref{fig: fig5}, shows that M4 has negative y coordinates, meaning that at least one subgroup was harmed while trying to achieve a smaller disparity between protected subgroups.

\section{Conclusion}
In this paper, we presented the notion of {\it positive-sum fairness} and argued that larger disparities are not necessarily harmful, as long as it does not come at the expense of a specific subgroup performance. The general performance, standard fairness and positive-sum fairness of four models was analyzed, each leveraging sensitive attributes in a different way. 

Our study highlights the need for a nuanced understanding of fairness metrics and their implications in real-world applications. Good incorporation of medical knowledge is crucial when utilizing sensitive information and evaluating fairness accurately, particularly in cases where models may show a large performance disparity. 

When traditional methods often aim for equality, positive-sum fairness focuses on equity, pushing for each group to achieve its highest possible performance level. This can lead to better overall outcomes, as it encourages to address the specific needs and challenges of each group without diminishing the quality of care for others. But, being defined as an optimization problem, it could also have unintended side effects as it may inadvertently prioritize larger or more well-represented groups by focusing the efforts on the groups with the highest impact on the overall performance rather than those with the most critical needs. Therefore, it is to be noted that meeting the positive-sum fairness criterion alone does not ensure a model to be fair from an egalitarian perspective, and the use of this notion in conjunction with other metrics can give a more holistic understanding of a model's fairness. 

As positive-sum fairness is a relative measure, it requires a baseline to be used. Further work in this area would include developing a more robust baseline or adapting the approach to remove the need for a baseline. It would also be worth it to compare out-of-domain tested models, include other sensitive attributes such as sex and age and take into account confounding factors.
\begin{credits}

\subsubsection{\discintname}
The authors declare that there are no conflicts of interest regarding the publication of this paper.
\end{credits}

\bibliographystyle{splncs04}
\bibliography{Paper-0006}

\end{document}